\documentclass{article} % For LaTeX2e
\usepackage{iclr2026_conference,times}

\usepackage{amsmath,amsfonts,bm}

\def\eqref#1{equation~\ref{#1}}
\def\1{\bm{1}}

\DeclareMathAlphabet{\mathsfit}{\encodingdefault}{\sfdefault}{m}{sl}
\SetMathAlphabet{\mathsfit}{bold}{\encodingdefault}{\sfdefault}{bx}{n}

\usepackage{hyperref}
\usepackage{url}

\usepackage{enumitem}
\usepackage{graphicx}
\usepackage{subcaption}
\usepackage{booktabs}
\usepackage{multirow}
\usepackage{makecell}
\newcommand{\partitle}[1]{\smallskip \noindent \textbf{#1.}}

\title{WaveMoE: A Wavelet-Enhanced Mixture-of-Experts Foundation Model for Time Series Forecasting}

\author{Shunyu Wu$^{1}$\thanks{Equal contribution.}, \ Jiawei Huang$^{1}$\footnotemark[1], \ Weibin Feng$^{1}$\footnotemark[1], \ Boxin Li$^{2}$, \ Xiao Zhang$^{2}$, \ Erli Meng$^{2}$, \\{\bfseries Dan Li$^{1}$\thanks{Dan Li is the Corresponding Author.},  \ Jian Lou$^{1}$, \ See-Kiong Ng$^{3}$} \\
$^{1}$Sun Yat-sen University \quad
$^{2}$Xiaomi Corporation \quad
$^{3}$National University of Singapore \\
\texttt{\{wushy88,huangjw255,fengwb6\}@mail2.sysu.edu.cn}, \\ 
\texttt{\{liboxin1,zhangxiao16,mengerli\}@xiaomi.com}, \\ 
\texttt{\{lidan263,louj5\}@mail.sysu.edu.cn}, \quad \texttt{seekiong@nus.edu.sg}
}

\iclrfinalcopy

\track{Research}
\begin{document}

\maketitle

\begin{abstract}
Time series foundation models (TSFMs) have recently achieved remarkable success in universal forecasting by leveraging large-scale pretraining on diverse time series data. Complementing this progress, incorporating frequency-domain information yields promising performance in enhancing the modeling of complex temporal patterns, such as periodicity and localized high-frequency dynamics, which are prevalent in real-world time series. To advance this direction, we propose a new perspective that integrates explicit frequency-domain representations into scalable foundation models, and introduce \textbf{WaveMoE}, a wavelet-enhanced mixture-of-experts foundation model for time series forecasting. WaveMoE adopts a dual-path architecture that jointly processes time series tokens and wavelet tokens aligned along a unified temporal axis, and coordinates them through a shared expert routing mechanism that enables consistent expert specialization while efficiently scaling model capacity. Preliminary experimental results on 16 diverse benchmark datasets indicate that WaveMoE has the potential to further improve forecasting performance by incorporating wavelet-domain corpora.
\end{abstract}

\section{Introduction}
\label{intro}

Time series data are ubiquitous in a wide range of real-world domains, including energy systems~\citep{tian2025new}, finance~\citep{kabir2025lstm}, healthcare~\citep{avinash2025time}, and industrial monitoring~\citep{saheed2025ga}. Time series forecasting, as a key task for dynamic-analysis and decisions, has traditionally been dominated by statistical methods~\citep{hyndman2018forecasting} and deep learning models trained in a domain-specific manner~\citep{nie2023time,xu2020tensorized,oreshkin2019n}. In recent years, propelled by the success of large language models (LLMs), time series foundation models (TSFMs) have emerged as a promising paradigm for universal forecasting through large-scale pretraining on multi-domain datasets~\citep{yao2025towards}. These models exhibit encouraging zero-shot and few-shot generalization capabilities, and in many cases match or even surpass the performance of task-specific forecasting models~\citep{aksu2024gift,liang2024foundation}.

Along this line of research, pioneering representative works include but are not limited to Chronos~\citep{ansari2024chronos}, MOMENT~\citep{goswami2024moment}, Timer~\citep{liu2024timer}, Moirai~\citep{woo2024unified}, Toto~\citep{cohen2024toto}, TimesFM~\citep{das2024decoder}, and so on, which adopt Transformer-based architectures and a large-scale pretraining paradigm to capture temporal dependencies for enhanced and generalizable forecasting performance. In addition, some models explore alternative architectures: Sundial~\citep{liu2025sundial} integrates diffusion-based modeling, TiRex~\citep{auer2025tirex} leverages LSTM-based architectures, and TabPFN-TS~\citep{hoo2025tables} employs a Tabular Prior-Data Fitted Network (PFN) model. More recent developments, including Chronos-2~\citep{ansari2025chronos}, Moirai 2.0~\citep{liu2025moirai}, TimesFM 2.5, FlowState~\citep{graf2025flowstate}, and YingLong~\citep{wang2025output} continue to advance TSFMs by systematically scaling data and model capacity while refining training strategies, collectively pushing the boundaries of robustness and cross-domain generalization. Building upon the scaling perspective, Time-MoE~\citep{xiaoming2025time} and MoiraiMoE~\citep{liu2025moiraimoe} further introduce a sparse mixture-of-experts architecture, enabling efficient scaling of TSFMs up to billions of parameters and achieving substantial gains in predictive accuracy. 

% In parallel, frequency-aware modeling has long been recognized as a complementary and informative perspective for time series analysis. Prior studies such as WaveToken~\citep{masserano2025enhancing} and WaveTS~\citep{zhou2025multi} demonstrate that incorporating wavelet-based representations can enrich temporal modeling by capturing periodic patterns, localized oscillations, and multi-scale dynamics commonly observed in real-world systems. Evidence further suggests that wavelet transformations can effectively preserve structured time--frequency information, providing useful inductive biases for learning complex temporal signals~\citep{yang2023waveform}.

Most existing TSFM studies focus on scaling the size of time-series corpora exclusively in the raw temporal domain. In this paper, we investigate an intriguing and less-explored \textbf{Research Question}: 

\textit{~~~~Whether scaling the pretraining corpora in the frequency domain, such as the wavelet domain, can further enhance the performance of time series foundation models (TSFMs)?}

Our motivation is that the frequency domains (e.g., wavelet domain) have long been recognized as a complementary and informative perspective for time-series analysis. Prior studies that incorporate frequency-domain representations into small-scale models, such as MLP-based architectures, have already demonstrated promising gains, suggesting that frequency-domain scaling may offer additional benefits at the foundation-model level. For example, WaveToken~\citep{masserano2025enhancing} and WaveTS~\citep{zhou2025multi} demonstrate that incorporating wavelet-based representations can enrich temporal modeling by capturing periodic patterns, localized oscillations, and multi-scale dynamics commonly observed in real-world systems. Evidence further suggests that wavelet transformations can effectively preserve structured time--frequency information, providing useful inductive biases for learning complex temporal signals~\citep{yang2023waveform}. 

% Existing works have largely focused on pretraining corpus solely from the raw temporal domain. In this paper, we investigate a new perspective on integrating wavelet-based frequency representations into scalable TSFMs. In this work, we propose \textbf{WaveMoE}, a wavelet-enhanced mixture-of-experts foundation model for time series forecasting. 
In this work, we address this motivating research question by proposing WaveMoE, a wavelet-enhanced mixture-of-experts foundation model for time series forecasting that instantiates wavelet-domain modeling on top of the popular MoE architecture. 
We propose a novel MoE block that accommodates both temporal and wavelet information pathways, forming a dual-path architecture for WaveMoE. Specifically, the frequency pathway leverages discrete wavelet transforms (DWT) for multi-scale decomposition to generate wavelet tokens, providing inductive bias for localized and oscillatory patterns~\citep{shensa2002discrete} while maintaining alignment with original time series tokens. A shared routing network under the MoE architecture performs unified token-to-expert routing across both pathways, coordinating time and frequency representations while enabling efficient capacity scaling~\citep{fedus2022switch}. To further reduce computational overhead, we introduce a sparse attention mechanism by activating tokens with top-$k$ attention scores~\citep{ma2025timeexpert}. At the output stage, independent prediction heads on the two pathways respectively forecast numerical values and corresponding wavelet components, and are jointly supervised by forecasting losses from both domains.
Preliminary experimental results on a diverse set of benchmark datasets indicate that WaveMoE has the potential to further improve forecasting performance by incorporating wavelet-domain corpora, thereby providing affirmative evidence for the research question. 

\partitle{Summary of Contributions} In summary, our main contributions are as follows:
\begin{itemize}[leftmargin=*]
    \item We investigate the less-explored research question of whether wavelet-based frequency-domain corpora can further enhance the forecasting performance of time series foundation models.
    \item We propose WaveMoE, a wavelet-enhanced mixture-of-experts foundation model that jointly models time-domain and frequency-domain representations. It employs a shared routing network for unified token-to-expert assignment, effectively coordinating time–frequency information while enabling scalable capacity.
    \item We provide preliminary empirical evidence across a wide range of benchmark datasets showing that WaveMoE can further enhance TSFM performance by leveraging wavelet-domain corpora.
\end{itemize}

\partitle{Outline of Appendix}
Due to the space limit, we defer the following content to Appendix:
\begin{itemize}[leftmargin=*]
    % \item Appendix~\ref{rw} contains extended related work that introduces time series foundation models and learning time series representations in frequency domains;
    \item Appendix~\ref{sec:pretraining_data} details the large-scale pretraining data construction process for WaveMoE;
    \item Appendix~\ref{sec:additional_results} reports additional benchmark results and qualitative visualization analyses to further examine the empirical behavior of WaveMoE.
\end{itemize}

\section{Related Work}
\label{rw}

\partitle{Time Series Foundation Models}
Time series foundation models (TSFMs) have recently transformed the landscape of time series forecasting by enabling versatile and powerful modeling across diverse temporal domains~\citep{liang2025foundation,liu2026unified}. Early research investigated adapting pretrained LLMs to time series tasks~\citep{jin2024time,gruver2023large}, while more recent approaches focus on pretraining large-scale models directly on extensive time series corpora~\citep{gong2025bridging,deng2026oats}, drawing inspiration from successful LLM architectures~\citep{das2024decoder}. This progress has fostered a rich ecosystem of Transformer-based TSFMs, including encoder-only~\citep{ansari2025chronos}, decoder-only~\citep{liu2025moirai,liu2025sundial}, and hybrid encoder–decoder~\citep{feng2025kairos} designs. Alongside architectural innovations, efficient scaling strategies such as mixture-of-experts (MoE) layers~\citep{xiaoming2025time,liu2025moiraimoe} have gained prominence by enabling large parameter counts with manageable computational cost, further boosting forecasting capabilities. While these models predominantly emphasize time-domain representations, exploring complementary frequency-domain information remains a promising avenue to enrich modeling. Our work contributes to this growing direction by proposing WaveMoE, a decoder-only MoE foundation model that integrates wavelet-based frequency features with time-domain features through coordinated dual pathways.

\partitle{Learning Time Series Representations in Frequency Domains}
Learning frequency-domain representations plays a vital role in capturing latent periodic patterns and high-frequency dynamics in time series, and has been widely studied in task-specific models~\citep{kim2025comprehensive,wang2024rose,lu2026towards}. Models such as Autoformer~\citep{wu2021autoformer} and FEDformer~\citep{zhou2022fedformer} incorporate Fourier transform features alongside time-domain signals, while WaveTS~\citep{zhou2025multi} and WaveForM~\citep{yang2023waveform} leverage discrete wavelet transforms (DWT) to achieve multi-scale, temporally localized decompositions. These approaches demonstrate the benefit of frequency-aware modeling to enhance prediction accuracy and interpretability. Recent efforts have also explored integrating wavelet tokenization within large foundation models to capture coarse-to-fine frequency structures~\citep{masserano2025enhancing}. Building on this foundation, WaveMoE leverages wavelet-based frequency representations to capture rich frequency structures, which are jointly coordinated with time-domain tokens via a shared MoE routing mechanism. This design seamlessly combines the strengths of both frequency and time domain modeling, making WaveMoE well-suited for scalable foundation models.

\begin{figure*}[!htbp]
    \centering
    \includegraphics[width=1.0\textwidth]{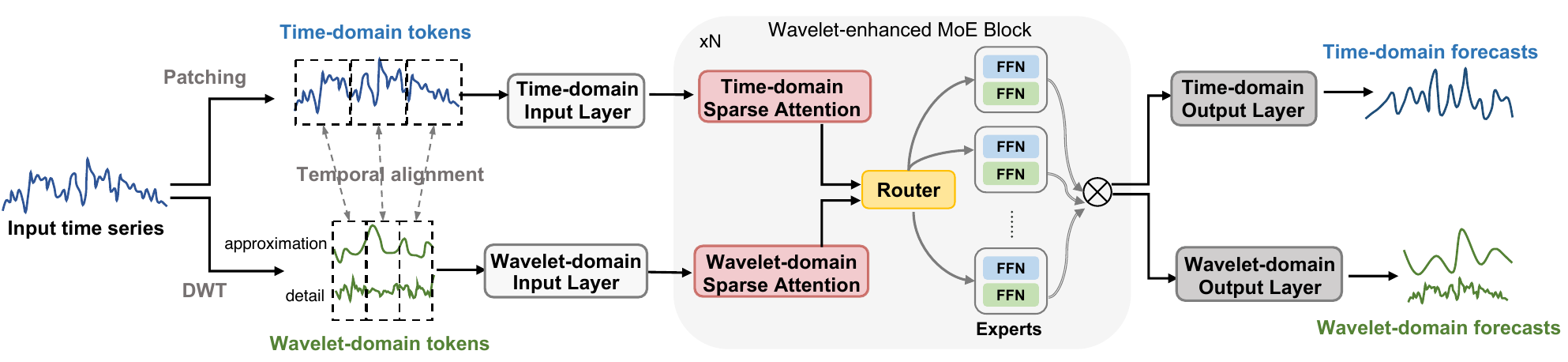}
    % \vskip -0.5em
    \caption{Overview of the proposed WaveMoE model.}
    \label{fig:framework}
    % \vspace{-1em}
\end{figure*}

\section{Methodology}

% \subsection{Overview of WaveMoE}

We propose WaveMoE, a wavelet-enhanced mixture-of-experts foundation model for time series forecasting. Given an input sequence $\mathbf{x}_{1:T}$, the objective is to predict a future horizon $\mathbf{x}_{T+1:T+H}$. For multivariate time series, we adopt a channel-independent strategy that decomposes multivariate inputs into univariate series, enabling scalable and flexible modeling across dimensions~\citep{nie2023time}.

\partitle{Overview of WaveMoE}
As illustrated in Figure~\ref{fig:framework}, WaveMoE introduces a novel design of the MoE block consisting of a time-domain pathway and a wavelet-domain pathway. Both pathways process the same input time series patch in parallel and produce temporally aligned token representations along a shared time axis. Within each pathway, a sparse attention module selectively aggregates informative tokens according to domain-specific characteristics. The resulting representations are then coordinated through a unified MoE routing mechanism, which jointly determines expert assignments for time and frequency domain tokens corresponding to the same temporal locations.  
At the output stage, independent prediction heads generate forecasts respectively in the original value space and the wavelet coefficient space, with joint supervision applied during training.

\partitle{Dual-Path Tokenization and Temporal Alignment}
WaveMoE begins by transforming the input time series into two parallel token sequences corresponding to the time and frequency domain pathways.
In the time-domain pathway, the input sequence is segmented into fixed-length patches, each aggregating a contiguous segment of time steps.
In the frequency-domain pathway, we apply a discrete wavelet transform (DWT) to decompose the input sequence into multi-scale approximation and detail wavelet coefficients, each retaining localized temporal information~\citep{liu2024disentangling}. To ensure compatibility with the time-domain pathway, wavelet coefficients are grouped into patches that are temporally aligned with their corresponding time-domain patches. Consequently, tokens from both pathways share a unified temporal indexing scheme, enabling consistent cross-domain coordination throughout the model.

\partitle{Sparse Attention over Token Sequences}
Both pathways employ independent Transformer-style self-attention layers to model dependencies among tokens. To reduce computational overhead and improve scalability to long input sequences, WaveMoE incorporates a sparse attention mechanism that selectively activates a subset of informative tokens~\citep{fedus2022switch}. Specifically, tokens with the top-$k$ attention scores are retained for interaction, while less informative tokens are masked.

\partitle{Unified Mixture-of-Experts Routing}
At the core of WaveMoE lies a unified mixture-of-experts (MoE) module that coordinates representation learning across the time and frequency domain pathways while scaling model capacity efficiently. Instead of employing separate routing mechanisms for each domain, WaveMoE adopts a shared routing strategy, where tokens from both domains, corresponding to identical temporal positions, are concatenated and assigned to experts in a unified manner.
Concretely, the routing network is implemented as a multi-layer perceptron (MLP) gating module, which dynamically generates routing scores over a set of experts based on the fused token representations. Each expert is equipped with a dual-branch feed-forward network to preserve domain-specific transformations: one feed-forward branch processes time-domain tokens, while the other processes wavelet-domain tokens. This design decouples time and frequency domain modeling within each expert, ensuring that domain-specific inductive biases are retained while maintaining structural consistency through shared routing~\citep{liu2025mofe}.

\section{Experiments}

\partitle{Model Training}
WaveMoE is pretrained on a large-scale time series corpus based on Time-300B~\citep{xiaoming2025time}, augmented with additional IoT datasets~\citep{liu2024timer} to improve real-world coverage. Details of the pretraining data construction are provided in Appendix~\ref{sec:pretraining_data}. The model is trained using the AdamW optimizer with a base learning rate of $2\times10^{-4}$ and a batch size of 128. Training is conducted for 100{,}000 steps with a warmup ratio of 0.1. The forecasting objective is optimized using the Huber loss~\citep{wen2019robusttrend}. In the frequency-domain pathway, WaveMoE utilizes the \texttt{bior2.2} wavelet as the basis function with a decomposition level of 2. The internal hyperparameters of WaveMoE are summarized in Table~\ref{tab:wavemoe-config}.

\begin{table}[!htbp]
% \vskip -1.0em
\centering
\caption{Internal configurations of WaveMoE.}
\vskip -1.0em
\label{tab:wavemoe-config}
\footnotesize
\begin{tabular}{c c c c c c c c c c}
\toprule
Layers &
Heads &
Experts &
\makecell[c]{Routing\\ Experts} &
\makecell[c]{Hidden\\ Size} &
\makecell[c]{FFN\\ Dim.} &
\makecell[c]{Patch\\ Length} &
\makecell[c]{Top-$k$\\ Attention} &
\makecell[c]{Activated\\ Params} &
\makecell[c]{Total\\ Params} \\
\midrule
12 & 12 & 8 & 2 & 384 & 1536 & 8 & 10 & 100M & 226M \\
\bottomrule
\end{tabular}
% \vspace{-0.3em}
\end{table}

\partitle{Experimental Settings}
We conduct a preliminary evaluation of WaveMoE on a diverse suite of 16 well-established benchmark datasets. WaveMoE is primarily compared with Time-MoE~\citep{xiaoming2025time}, which shares a similar MoE architecture, to assess the effectiveness of the proposed wavelet tokenization design. In addition, earlier representative TSFMs including Timer~\citep{liu2024timer}, Chronos~\citep{ansari2024chronos}, and Sundial~\citep{liu2025sundial}, are included as reference baselines. In all experiments, the context length is fixed at 512 time steps, and the prediction horizon is set to 96. The evaluation metrics are Mean Squared Error (MSE) and Mean Absolute Error (MAE).

\partitle{Main Results}
As summarized in Table~\ref{tab:main-results}, WaveMoE demonstrates significant advantages on the majority of datasets. Specifically, it achieves the best MSE scores on 14 out of 16 datasets and the best MAE scores on 11 datasets, demonstrating clear and consistent improvements over the competing baselines. Additionally, WaveMoE attains the lowest average MSE and MAE values overall, highlighting its superior stability and robustness. These results indicate that WaveMoE not only excels in forecasting accuracy but also generalizes well across a variety of time series domains, confirming its effectiveness as a versatile and reliable forecasting model.
For a more comprehensive evaluation, we provide extended benchmark results in Appendix~\ref{subsec:extended_benchmarks} and detailed visualization analyses in Appendix~\ref{subsec:visualization_analysis}.

\begin{table}[h]
% \vskip -0.5em
\centering
\caption{Forecasting performance (MSE and MAE) of WaveMoE compared with baseline models on 16 benchmark datasets. The best results for each dataset are \textbf{bolded}.}
\label{tab:main-results}
\vskip -1.0em
\resizebox{\linewidth}{!}
{
\begin{tabular}{l c c c c c c c c c c}
\toprule
\multirow{2}{*}{Dataset} & \multicolumn{2}{c}{WaveMoE} & \multicolumn{2}{c}{Time-MoE} & \multicolumn{2}{c}{Timer} & \multicolumn{2}{c}{Chronos} & \multicolumn{2}{c}{Sundial} \\
 & MSE & MAE & MSE & MAE & MSE & MAE & MSE & MAE & MSE & MAE \\
\midrule
ETT1 & 0.2012 & 0.2578 & \textbf{0.1756} & 0.2498 & 0.9760 & 0.5020 & 0.4456 & 0.3540 & 0.2052 & \textbf{0.2403} \\
ETT2 & 0.0342 & 0.1226 & 0.0840 & 0.1809 & 0.0361 & 0.1250 & 0.0357 & 0.1200 & \textbf{0.0245} & \textbf{0.0958} \\
Exchange Rate & \textbf{0.0284} & \textbf{0.1050} & 0.0320 & 0.1168 & 0.0416 & 0.1158 & 0.0353 & 0.1119 & 0.0382 & 0.1139 \\
M1 Monthly & \textbf{0.0009} & 0.0232 & 0.0060 & 0.0704 & 0.0909 & 0.2219 & 0.0869 & \textbf{0.0192} & 0.1816 & 0.0371 \\
M1 Quarterly & \textbf{0.0012} & 0.0291 & 0.0052 & 0.0667 & 0.0073 & 0.1923 & 0.0279 & \textbf{0.0089} & 0.1168 & 0.0234 \\
M1 Yearly & \textbf{0.0009} & 0.0254 & 0.0029 & 0.0493 & 0.0502 & 0.1684 & 0.0489 & \textbf{0.0070} & 0.1152 & 0.0166 \\
M5 & \textbf{0.0688} & \textbf{0.0150} & 0.8408 & 0.4251 & 1.2633 & 0.6218 & 1.0564 & 0.3751 & 0.9121 & 0.3717 \\
Monash M3 & \textbf{0.0003} & \textbf{0.0109} & 0.0077 & 0.0768 & 0.1187 & 0.3093 & 0.1223 & 0.0321 & 0.1392 & 0.0460 \\
NN5 & \textbf{0.0002} & \textbf{0.0101} & 0.0080 & 0.0785 & 0.1552 & 0.3727 & 0.1210 & 0.0461 & 0.2371 & 0.0808 \\
Traffic & \textbf{0.1030} & \textbf{0.0618} & 0.1827 & 0.2145 & 2.0162 & 1.0998 & 0.5507 & 0.3475 & 0.2294 & 0.2095 \\
Weather & \textbf{0.1553} & \textbf{0.0471} & 0.5984 & 0.4396 & 0.9986 & 0.6532 & 0.7963 & 0.4532 & 0.6350 & 0.4066 \\
M4 Monthly & \textbf{0.2500} & \textbf{0.2559} & 0.3736 & 0.3017 & 0.2750 & 0.3017 & 0.2709 & 0.2807 & 0.3232 & 0.3215 \\
Entsoe & \textbf{0.2373} & \textbf{0.3438} & 0.2832 & 0.3811 & 0.9728 & 0.8412 & 1.0461 & 0.7691 & 0.3375 & 0.3945 \\
Solar with Weather & \textbf{0.4355} & \textbf{0.3359} & 0.6147 & 0.4435 & 1.8617 & 0.9086 & 1.5574 & 0.6519 & 0.5978 & 0.3741 \\
UK Covid & \textbf{1.1719} & \textbf{0.6484} & 1.5570 & 0.7433 & 1.4990 & 0.8294 & 1.8201 & 0.8303 & 1.5359 & 0.8158 \\
Sensor Data & \textbf{0.4181} & \textbf{0.4863} & 0.4958 & 0.5307 & 0.9514 & 0.7651 & 1.5138 & 0.8923 & 0.7164 & 0.6348 \\
\midrule
Average & \textbf{0.1942} & \textbf{0.1736} & 0.3292 & 0.2730 & 0.7071 & 0.5018 & 0.5960 & 0.3312 & 0.3966 & 0.2614 \\
\midrule
\multicolumn{1}{l}{\# Best} & 14 & 11 & 1 & 0 & 0 & 0 & 0 & 3 & 1 & 2 \\
\bottomrule
\end{tabular}
}
\end{table}

\section{Concluding Remarks and Future Work}
% We propose WaveMoE, a mixture-of-experts time series foundation model that explicitly integrates time-domain and wavelet-based frequency-domain representations within a unified and scalable architecture. Through a dual-path design and a shared MoE routing mechanism, WaveMoE effectively coordinates time--frequency information while enabling efficient capacity scaling. Extensive experiments on 16 benchmark datasets demonstrate that WaveMoE consistently outperforms strong TSFM baselines, highlighting its effectiveness and generalization capability. Future work may include extending WaveMoE to model multivariate dependencies, exploring more adaptive time--frequency fusion mechanisms, and developing time--frequency aligned interpretability analyses to better understand forecasting dynamics.

We propose WaveMoE, a mixture-of-experts time series foundation model that explicitly integrates time-domain and wavelet-based frequency-domain representations within a unified and scalable architecture. Through its dual-path design and shared MoE routing mechanism, WaveMoE provides preliminary evidence that coordinating time–frequency information can be beneficial for forecasting. Current experiments on 16 benchmark datasets suggest that the wavelet tokenization approach improves performance over baseline TSFMs in certain settings, though further evaluation is needed. Future work includes extending WaveMoE to model multivariate dependencies, exploring more adaptive time–frequency fusion strategies, and developing interpretability analyses aligned with time–frequency representations to better understand and leverage forecasting dynamics.


\section*{Acknowledgments}
We would like to sincerely thank all anonymous reviewers for their valuable feedback and constructive comments in improving the quality of our paper. We are also grateful to the workshop organizers for hosting this inspiring event and providing a valuable platform for discussions within the TSALM community.
Finally, we thank our lab colleagues Zhuomin Chen, Jiahui Zhou, Xiangting Wu, and Haozheng Ye for their helpful discussions and assistance with pretraining data preparation during the development of this work.

\bibliography{iclr2026_conference}
\bibliographystyle{iclr2026_conference}

\clearpage
\appendix

\section{Pretraining Data Construction}
\label{sec:pretraining_data}

\subsection{Dataset Composition and Domain Coverage}
WaveMoE is pretrained on a large-scale dataset built upon the Time-300B corpus~\citep{xiaoming2025time}. Specifically, the pretraining data are derived from the publicly available Time-300B dataset after domain balancing and multi-stage quality filtering, further augmented with additional Internet of Things (IoT) data from Unified Time Series Dataset (UTSD)~\citep{liu2024timer}.

Time-300B covers nine major domains and comprises over 300 billion time steps in total. However, the original dataset exhibits substantial domain imbalance, with certain scenarios heavily overrepresented. To address this issue, we perform domain-aware filtering to reduce redundancy in dominant scenarios while preserving representative temporal patterns across domains. In addition, IoT data from USTD are incorporated to further enhance domain diversity. 

After domain balancing, filtering, and preprocessing, the final pretraining dataset contains approximately 98 billion time steps. It spans 9 domains, including IoT, energy, finance, healthcare, nature, sales, transportation, network systems and synthetic data. The resulting dataset preserves multi-scale sampling frequencies while removing low-quality and severely corrupted samples, thereby ensuring both temporal diversity and data reliability and providing a solid foundation for large-scale pretraining.

\subsection{Data Preprocessing Pipeline}
To ensure data quality and consistency during pretraining, WaveMoE adopts a unified preprocessing pipeline that includes window segmentation, quality filtering, missing-value handling, and balanced sampling.

\partitle{Window Segmentation Strategy}
To accommodate the fixed input length requirement of Transformer-based architectures, raw time series are segmented into fixed-length training samples using a sliding window mechanism. The window length is uniformly set to 4096 time steps, which is sufficient to capture long-range temporal dependencies while maintaining computational efficiency. 
For sequences with length greater than or equal to 4096, non-overlapping sliding windows (stride = 4096) are applied. This design ensures sample completeness while avoiding redundancy caused by overlapping windows, thereby improving training efficiency. 
For shorter sequences with length less than 4096, we adopt a sequence packing strategy instead of zero-padding. Multiple short sequence fragments are concatenated to form a full-length window, maximizing data utilization and reducing unnecessary padding. Compared to conventional padding-based approaches, sequence packing significantly improves effective data usage.

\partitle{Quality Filtering}
To mitigate the influence of low-quality data on model training, a multi-stage quality filtering mechanism is applied at the window level. First, we compute the proportion of missing values (NaN or Inf) within each window. Windows with more than 20\% missing entries are discarded. This check is performed prior to normalization to ensure evaluation is based on raw data quality.
Second, we examine the proportion of zero or near-zero values (absolute value less than $1 \times 10^{-6}$). If such values exceed 20\% of the window, the sample is considered invalid, as it may correspond to prolonged inactivity or sensor malfunction.
Third, we assess sequence variability using first- and second-order differences. If the proportion of zero or near-zero values in either the first- or second-order difference exceeds 20\%, the sequence is regarded as overly smooth and lacking informative dynamics, and is therefore removed. 
These filtering criteria are consistent with practices adopted in Time-MoE~\citep{xiaoming2025time} and ensure that retained windows exhibit sufficient temporal variation, preventing the model from learning trivial or static patterns.

\partitle{Missing-Value Handling}
For windows that pass quality filtering, a unified missing-value handling strategy is applied. All NaN and Inf values are replaced with zero to ensure numerical stability. In addition, a corresponding loss mask is generated for each window to indicate valid positions. Locations originally containing NaN or Inf are assigned a mask value of 0 and excluded from loss computation. Original zero-valued positions are also masked out to avoid potential interference during training. All other valid positions are assigned a mask value of 1 and participate normally in loss calculation. 
This mechanism ensures that the model learns exclusively from reliable and informative observations while preventing missing or anomalous values from adversely affecting training.

\partitle{Balanced Sampling}
To improve cross-domain generalization, each window is annotated with its domain or dataset identifier during preprocessing to enable balanced sampling at training time. Since the pretraining corpus spans multiple domains with highly uneven data distributions, naive random sampling would bias the model toward data-rich domains while underrepresenting smaller ones. 
By adopting a balanced sampling strategy, each training batch draws samples uniformly from different data subsets, ensuring that the model learns diverse temporal patterns across domains. This design enhances the robustness and generalization ability of WaveMoE in heterogeneous real-world forecasting scenarios.

\section{Additional Experimental Results and Analyses}
\label{sec:additional_results}

\subsection{Extended Benchmark Evaluation}
\label{subsec:extended_benchmarks}

To provide a more comprehensive evaluation of WaveMoE, we further report forecasting results on a broader set of benchmark datasets, as summarized in Table~\ref{tab:additional-results}. These datasets cover diverse real-world scenarios, including electricity demand forecasting, tourism statistics, traffic flow prediction, hierarchical retail sales, public health records, and large-scale transactional data, thereby offering a more extensive assessment of cross-domain generalization.

All experiments follow the same configuration as in the main results, with a fixed context length of 512 and a prediction horizon of 96. Performance is evaluated using Mean Squared Error (MSE) and Mean Absolute Error (MAE) for consistent comparison across models. Overall, the results on these additional benchmarks are consistent with the main experimental findings. WaveMoE achieves competitive or leading performance across the majority of datasets and maintains strong stability under diverse temporal patterns and data distributions.

\begin{table}[h]
\centering
\caption{Forecasting performance (MSE and MAE) of WaveMoE compared with baseline models on additional benchmark datasets. The best results for each dataset are \textbf{bolded}.}
\label{tab:additional-results}
\resizebox{\linewidth}{!}{
\begin{tabular}{l c c c c c c c c c c}
\toprule
\multirow{2}{*}{Dataset} & \multicolumn{2}{c}{WaveMoE} & \multicolumn{2}{c}{Time-MoE} & \multicolumn{2}{c}{Timer} & \multicolumn{2}{c}{Chronos} & \multicolumn{2}{c}{Sundial} \\
 & MSE & MAE & MSE & MAE & MSE & MAE & MSE & MAE & MSE & MAE \\
\midrule
Australian Electricity  & \textbf{0.1454} & \textbf{0.2438} & 0.1508 & 0.2480 & 1.8303 & 1.1054 & 0.6366 & 0.5632 & 0.1926 & 0.2805 \\
CIF 2016 12 & \textbf{0.0002} & \textbf{0.0023} & 0.0069 & 0.0735 & 0.1253 & 0.3015 & 0.0259 & 0.0159 & 0.2634 & 0.0581 \\
CIF 2016 6 & \textbf{0.0003} & 0.0040 & 0.0041 & 0.0602 & 0.0761 & 0.1561 & 0.0017 & \textbf{0.0033} & 0.0014 & 0.0061 \\
ERCOT & \textbf{0.2080} & \textbf{0.3152} & 0.2085 & 0.3188 & 1.4126 & 0.9399 & 0.4388 & 0.4874 & 0.2641 & 0.3553 \\
Tourism Monthly & \textbf{0.0002} & \textbf{0.0023} & 0.0106 & 0.0886 & 0.2783 & 0.3609 & 0.0459 & 0.0167 & 0.0492 & 0.0271 \\
Tourism Quarterly & \textbf{0.0003} & \textbf{0.0037} & 0.0060 & 0.0704 & 0.0680 & 0.2047 & 0.0634 & 0.0183 & 0.0469 & 0.0147 \\
Tourism Yearly & \textbf{0.0003} & \textbf{0.0038} & 0.0029 & 0.0488 & 0.0171 & 0.0883 & 0.0632 & 0.0093 & 0.1153 & 0.0202 \\
Loop Seattle & 0.5493 & \textbf{0.4190} & \textbf{0.5051} & 0.4284 & 1.0182 & 0.6393 & 1.1660 & 0.6697 & 0.7220 & 0.5102 \\
M Dense D & \textbf{0.0256} & 0.0690 & 0.0332 & 0.1014 & 0.0694 & 0.1411 & 0.0447 & 0.0898 & 0.0270 & \textbf{0.0652} \\
M Dense H & 0.0500 & 0.0950 & \textbf{0.0494} & 0.1001 & 1.2013 & 0.6105 & 0.1352 & 0.1532 & 0.0509 & \textbf{0.0901} \\
SZ Taxi & \textbf{0.3893} & \textbf{0.3983} & 0.4058 & 0.4446 & 0.4414 & 0.4626 & 0.4610 & 0.4463 & 0.4490 & 0.4407 \\
Hierarchical Sales & \textbf{0.6344} & \textbf{0.4299} & 0.6898 & 0.4579 & 0.7006 & 0.4736 & 0.9651 & 0.5190 & 0.7649 & 0.4790 \\
Favorita Transactions & 0.1520 & 0.1980 & 0.2609 & 0.3451 & 0.1841 & 0.2940 & 0.1650 & 0.2239 & \textbf{0.1349} & \textbf{0.1768} \\
KDD Cup & 0.9916 & \textbf{0.6852} & \textbf{0.8341} & 0.7148 & 1.0314 & 0.8020 & 1.2360 & 0.7846 & 1.2326 & 0.8036 \\
Redset & \textbf{0.6015} & \textbf{0.2641} & 0.6124 & 0.2804 & 0.7433 & 0.3826 & 0.8618 & 0.3446 & 0.6578 & 0.2944 \\
Hospital Admissions & \textbf{1.0079} & \textbf{0.7986} & 1.0172 & 0.8032 & 1.0339 & 0.8119 & 1.2199 & 0.8713 & 1.3127 & 0.8957 \\
Bizitobs L2C & 0.1588 & 0.2460 & 0.1775 & 0.2633 & 0.4183 & 0.3806 & \textbf{0.1312} & \textbf{0.2042} & 0.4172 & 0.3653 \\
Boomlet & 0.8300 & \textbf{0.4458} & \textbf{0.7747} & 0.4662 & 0.8704 & 0.5251 & 1.1955 & 0.5616 & 0.8726 & 0.4722 \\
\bottomrule
\end{tabular}
}
\end{table}

\subsection{Visualization Analysis of WaveMoE Forecasting Results}
\label{subsec:visualization_analysis}

To complement the quantitative evaluation, we provide qualitative visualization comparisons between WaveMoE and representative baseline models. These visual analyses aim to illustrate how different models capture temporal dynamics, including peak–trough localization, oscillation amplitude, periodic structure, and abrupt trend transitions. Across diverse datasets, WaveMoE consistently demonstrates stronger fidelity to the ground-truth series, particularly under complex multi-scale and high-frequency temporal patterns.

\begin{figure*}[!htbp]
\centering
\begin{subfigure}{0.32\textwidth}
    \centering
    \includegraphics[width=\linewidth]{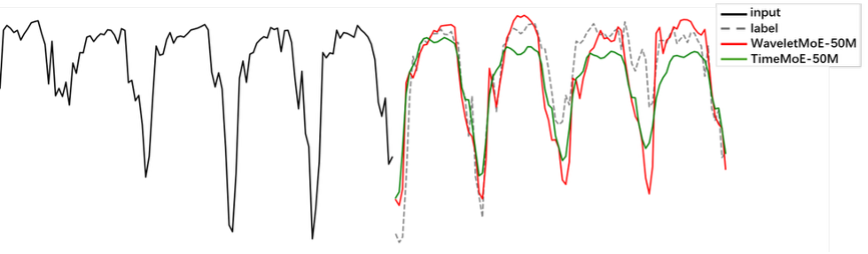}
    \caption{Ercot dataset}
    \label{fig:vis_ercot}
\end{subfigure}
\hfill
\begin{subfigure}{0.32\textwidth}
    \centering
    \includegraphics[width=\linewidth]{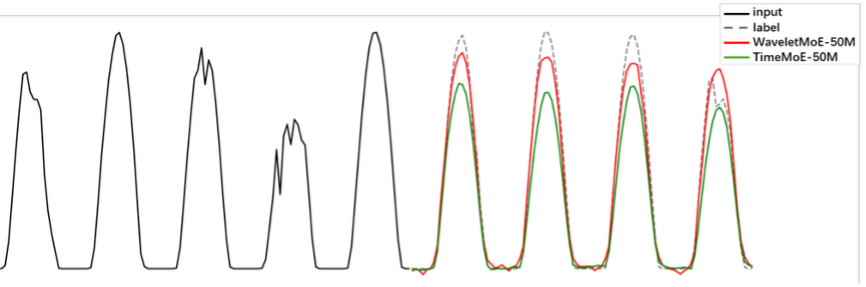}
    \caption{M Dense H dataset} 
    \label{fig:vis_mdense_h}
\end{subfigure}
\hfill
\begin{subfigure}{0.32\textwidth}
    \centering
    \includegraphics[width=\linewidth]{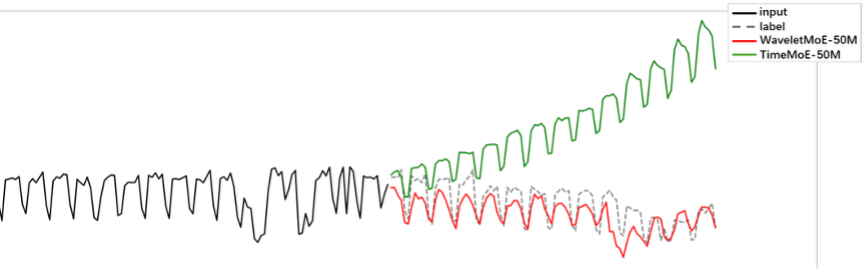}
    \caption{M Dense D dataset}
    \label{fig:vis_mdense_d}
\end{subfigure}
\caption{Forecast comparisons between WaveMoE and Time-MoE across representative datasets.}
\label{fig:vis_timemoe}
\end{figure*}

\partitle{Comparison Between WaveMoE and Time-MoE}
Figure~\ref{fig:vis_timemoe} shows forecast comparisons between WaveMoE and Time-MoE~\citep{xiaoming2025time} on example time series from three benchmark datasets. Overall, WaveMoE demonstrates closer alignment with the ground-truth series, particularly in terms of peak and trough localization, amplitude reconstruction, and trend consistency. While Time-MoE generally captures the overall trajectory, noticeable deviations remain around extreme values. In cases of rapid temporal transitions, such as sharp spikes, deep troughs, or frequent high-frequency oscillations, WaveMoE more accurately recovers both the amplitude and frequency characteristics, whereas Time-MoE tends to produce smoother forecasts that partially attenuate high-frequency fluctuations.

\begin{figure*}[!htbp]
\centering
\begin{subfigure}{0.32\textwidth}
    \centering
    \includegraphics[width=\linewidth]{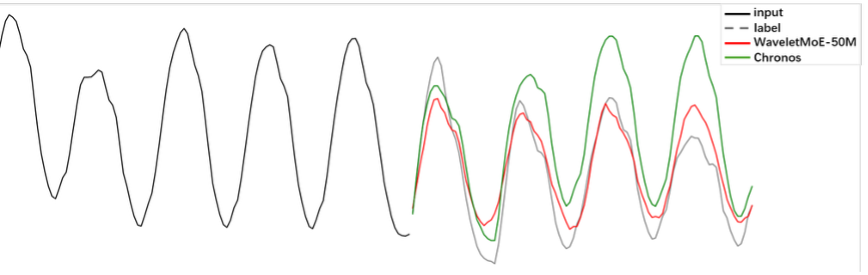}
    \caption{Ercot dataset}
    \label{fig:vis_ercot}
\end{subfigure}
\hfill
\begin{subfigure}{0.32\textwidth}
    \centering
    \includegraphics[width=\linewidth]{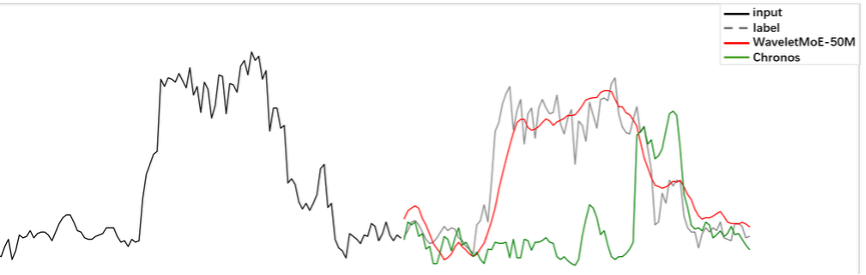}
    \caption{ETT1 dataset}
    \label{fig:vis_ett1}
\end{subfigure}
\hfill
\begin{subfigure}{0.32\textwidth}
    \centering
    \includegraphics[width=\linewidth]{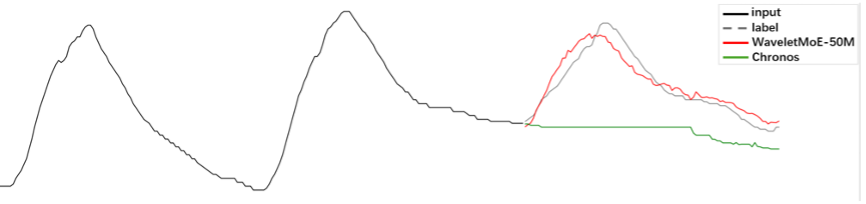}
    \caption{ETT2 dataset}
    \label{fig:vis_ett2}
\end{subfigure}
\caption{Forecast comparisons between WaveMoE and Chronos across representative datasets.}
\label{fig:vis_chronos}
\end{figure*}

\partitle{Comparison Between WaveMoE and Chronos}
Figure~\ref{fig:vis_chronos} visualizes forecast comparisons between WaveMoE and Chronos~\citep{ansari2024chronos}. In datasets exhibiting pronounced periodic patterns, WaveMoE more precisely captures peak and trough amplitudes as well as the underlying periodic rhythms. Chronos effectively models the global trend but often generates smoother trajectories, which can obscure fine-grained periodic structures. Under high-frequency fluctuation scenarios, WaveMoE closely follows rapid changes in the ground-truth sequence, whereas Chronos occasionally exhibits lag during abrupt directional shifts.

\begin{figure*}[!htbp]
\centering
\begin{subfigure}{0.32\textwidth}
    \centering
    \includegraphics[width=\linewidth]{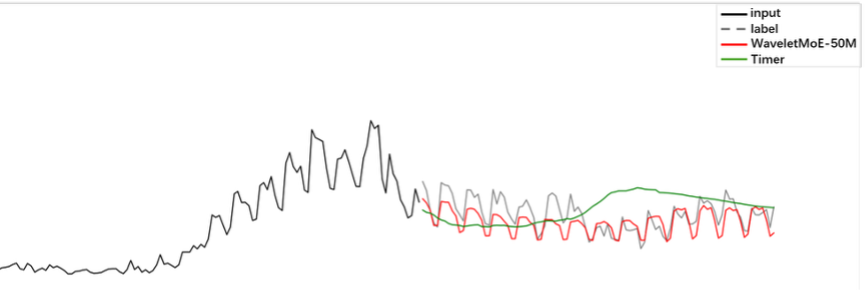}
    \caption{UK Covid dataset}
    \label{fig:vis_ukcovid}
\end{subfigure}
\hfill
\begin{subfigure}{0.32\textwidth}
    \centering
    \includegraphics[width=\linewidth]{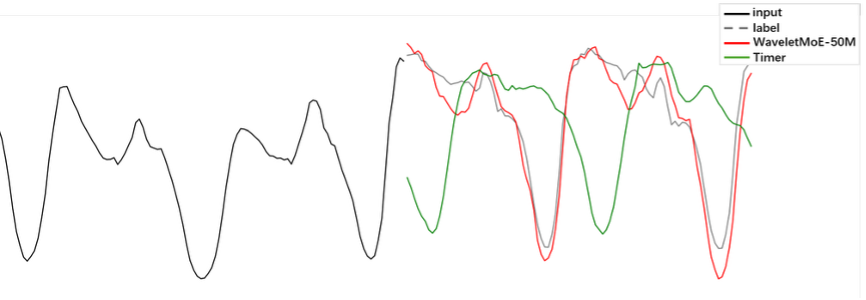}
    \caption{Australian Electricity dataset}
    \label{fig:vis_aus_elec}
\end{subfigure}
\hfill
\begin{subfigure}{0.32\textwidth}
    \centering
    \includegraphics[width=\linewidth]{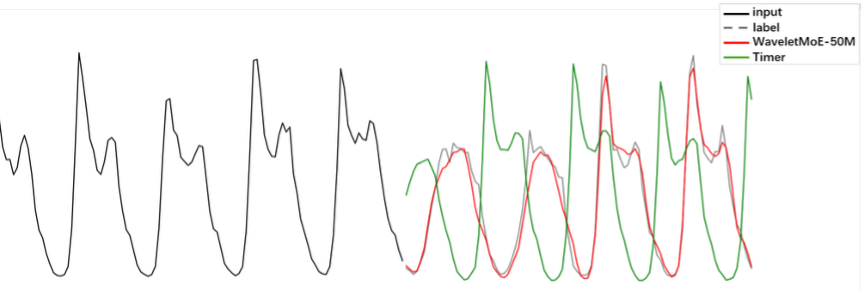}
    \caption{Traffic dataset}
    \label{fig:vis_traffic}
\end{subfigure}
\caption{Forecast comparisons between WaveMoE and Timer across representative datasets.}
\label{fig:vis_timer}
\end{figure*}

\partitle{Comparison Between WaveMoE and Timer}
Figure~\ref{fig:vis_timer} presents visualization comparisons between WaveMoE and Timer~\citep{liu2024timer}. WaveMoE captures both the overall trend and high-frequency fluctuations, maintaining alignment with the ground-truth series across peaks and troughs. Timer exhibits a stronger smoothing tendency, resulting in attenuation of fine-grained oscillatory components. Additionally, Timer shows delayed responses to abrupt trend changes, whereas WaveMoE adapts more rapidly to directional transitions and maintains sharper turning-point fidelity.

\end{document}